%% file: ms.tex
\documentclass[letterpaper, 10 pt, conference]{ieeeconf}  

\IEEEoverridecommandlockouts                              

\usepackage{graphics} 
\usepackage{epsfig} 
\usepackage{amsmath} 

\usepackage{xcolor}
\usepackage{stix}
\usepackage{tikz}

\newcommand{\tinytext}[1]{\ensuremath{\textnormal{\tiny{#1}}}}
\newcommand{\definedas}{\ensuremath{\stackrel{\textnormal{\tiny{def}}}{=}}}

\title{
	\LARGE \bf Urban Traffic Surveillance (UTS):
	\\
	A fully probabilistic 3D tracking approach based on 2D detections
}

\author{Henry Bradler$^{\dagger, 1}$, Adrian Kretz$^{\dagger, 1}$ and Rudolf Mester$^{2}$
\thanks{*This project (HA project no. 626/18-49) is financed with funds of LOEWE -- Landes-Offensive zur Entwicklung Wissenschaftlich-\"okonomischer Exzellenz, F\"orderlinie~3: KMU-Verbundvorhaben (State Offensive for the Development of Scientific and Economic Excellence).}
\thanks{$^{\dagger}$ Both authors contributed equally}
\thanks{$^{1}$VSI Lab, CS Dept., Goethe University, Frankfurt, Germany {\tt\small \{bradler,kretz\}@vsi.cs.uni-frankfurt.de}}%
\thanks{$^{2}$Norwegian Open AI Lab, CS Dept. (IDI), NTNU, Trondheim, Norway {\tt\small rudolf.mester@ntnu.no}}%
}

\newcommand\makecopyrightnotice{%
	\begin{tikzpicture}
		[remember picture,overlay]\node[anchor=south,yshift=10pt] at (current page.south)
		{\fbox{\parbox{\dimexpr\textwidth-\fboxsep-\fboxrule\relax}{
				\footnotesize © 2021 IEEE. Personal use of this material is permitted. Permission from IEEE must be obtained for all other uses, in any current or future media, including reprinting/republishing this material for advertising or promotional purposes, creating new collective works, for resale or redistribution to servers or lists, or reuse of any copyrighted component of this work in other works.
		}}};
	\end{tikzpicture}%
}

\begin{document}

\maketitle
\thispagestyle{empty}
\pagestyle{empty}


\input{sourcefiles/abstract.tex}

\input{sourcefiles/introduction.tex}
\input{sourcefiles/related-work.tex}
\input{sourcefiles/approach.tex}
\input{sourcefiles/experiments.tex}
\input{sourcefiles/conclusion.tex}


\addtolength{\textheight}{0cm}  


{
	\small
	\bibliographystyle{ieee}
	\bibliography{sourcefiles/bibliography.bib}
}


\end{document}

%% file: sourcefiles/abstract.tex
\begin{abstract}

\emph{Urban Traffic Surveillance} (UTS) is a surveillance system based on a monocular and calibrated video camera that detects vehicles in an urban traffic scenario with dense traffic on multiple lanes and vehicles performing sharp turning maneuvers. UTS then tracks the vehicles using a 3D bounding box representation and a physically reasonable 3D motion model relying on an unscented Kalman filter based approach. Since UTS recovers positions, shape and motion information in a three-dimensional world coordinate system, it can be employed to recognize diverse traffic violations or to supply intelligent vehicles with valuable traffic information. We build on YOLOv3 as a detector yielding 2D bounding boxes and class labels for each vehicle. A 2D detector renders our system much more independent to different camera perspectives as a variety of labeled training data is available. This allows for a good generalization while also being more hardware efficient. The task of 3D tracking based on 2D detections is supported by integrating class specific prior knowledge about the vehicle shape. We quantitatively evaluate UTS using self generated synthetic data and ground truth from the CARLA simulator, due to the non-existence of datasets with an urban vehicle surveillance setting and labeled 3D bounding boxes. Additionally, we give a qualitative impression of how UTS performs on real-world data. Our implementation is capable of operating in real time on a reasonably modern workstation. To the best of our knowledge, UTS is to date the only 3D vehicle tracking system in a surveillance scenario (static camera observing moving targets).

\end{abstract}

%% file: sourcefiles/introduction.tex
\section{Introduction}

\makecopyrightnotice
In this paper, we present a surveillance system called \emph{Urban Traffic Surveillance} (UTS) that is designed to work with monocular and fully, intrinsically and extrinsically calibrated video cameras that observe vehicles in an urban traffic scenario. We focus on a demanding scene composed of an intersection of roads with multiple lanes, dense traffic and possibilities to perform sharp turning maneuvers and U-turns. The goal is to recover 3D information for each of the vehicles passing the intersection. The 3D information consists of metric shapes, knowledge about the states of motion and trajectories of the positions of the vehicles given in a camera independent world coordinate system. By this, UTS can act as a smart infrastructure component providing intelligent vehicles or city planners with valuable information about the traffic flow.

The desired 3D information is achieved by deploying a 3D tracking approach with each car being represented by a three-dimensional bounding box whose motion obeys a physically reasonable motion model. In contrast to other 3D vehicle tracking approaches, UTS is specifically designed for surveillance applications, i.e. the cameras are fixed to the environment (e.g. traffic light poles, buildings) and are not located inside a car. As a consequence, this elevated position leads to a slightly down facing perspective. This is different to other similar approaches where the cameras are located inside of vehicles and the perspective is always chosen to be parallel to the street plane and all the other traffic participants.

Seemingly contradictory to our 3D approach, we rely on a YOLOv3~\cite{redmon2018CoRR} detector which only yields 2D detections. This is because today's state of the art 3D detectors based on deep networks (e.g. M3D-RPN~\cite{Brazil2019ICCV}) only poorly generalize to differing camera perspectives than the ones they were trained on. Using a 2D detector, on the other hand, has the advantage that a variety of diverse labeled training data is available and those detectors are well-engineered and generalize well to different perspectives while also being more hardware efficient.
\begin{figure}[t!]
	\centering
	\includegraphics[width = 0.3\linewidth]{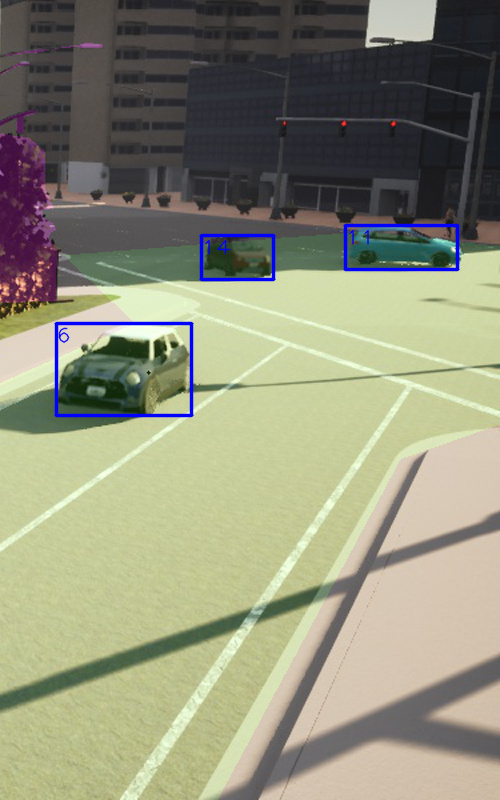}
	\includegraphics[width = 0.3\linewidth]{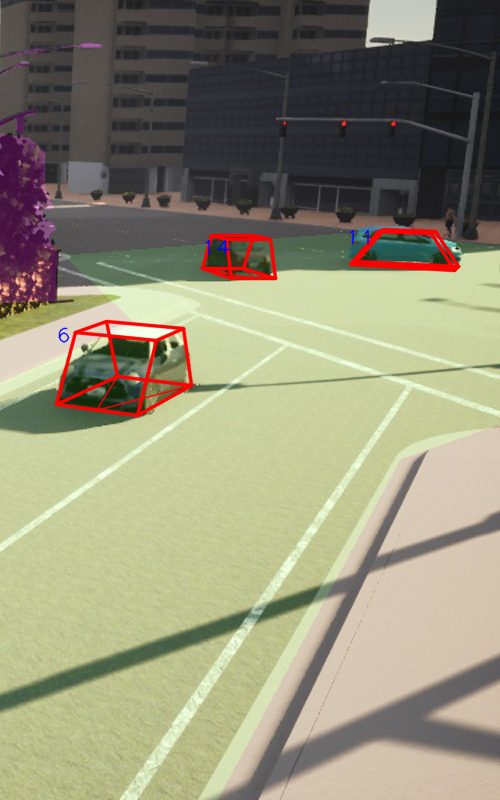}
	\includegraphics[width = 0.3\linewidth]{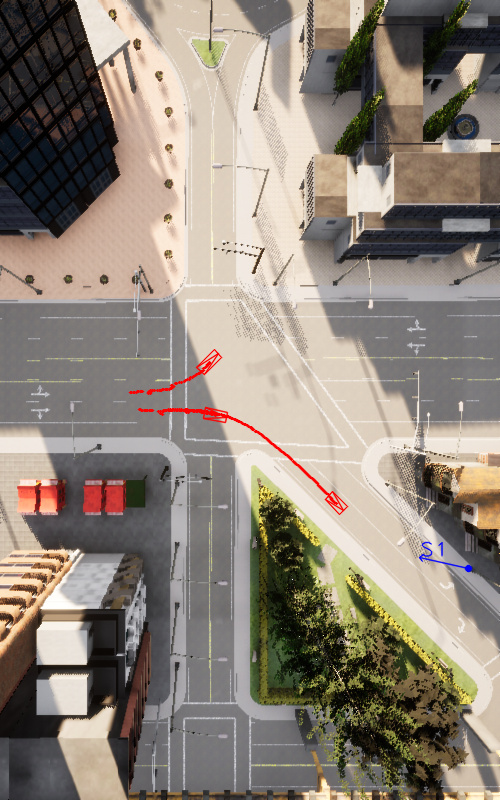}
	\caption{The left image shows the output of the detector (axis aligned 2D bounding boxes). The reconstructed and tracked 3D bounding boxes are depicted in the image in the middle. The area masked in green color in the first two images is the detection and tracking area that defines where our system should actively operate. The right image shows the same 3D bounding boxes and the center trajectories from a bird's eye view. Images are generated using CARLA Simulator~\cite{Dosovitskiy2017PMLR}. Best viewed in color.}
	\label{fig:teaser}
\end{figure}

The main contribution of this paper is to present a novel 3D tracking approach specifically designed for traffic surveillance in urban settings (see Fig. \ref{fig:teaser}). Due to the different perspective and the lack of training data with annotated 3D ground truth, we rely on 2D detections. We emphasize this aspect, as typically tracking and detection take place in the same dimension. Recovering tracks of 3D bounding boxes only using 2D detections is very challenging, as the mapping from 2D boxes to corresponding 3D boxes is underdetermined as opposed to the other direction. We overcome this challenging limitation by using a sophisticated 3D initialization component, a realistic 3D motion model and an unscented Kalman filter approach that can tackle the strong nonlinearities of the observation model. To the best of our knowledge UTS is the only approach to perform 3D reconstruction out of 2D observations in an urban surveillance scenario with static cameras.

%% file: sourcefiles/related-work.tex
\section{Related Work}

We estimate initial 3D boxes containing the detected vehicles by first estimating the
direction of movement and subsequently fitting 3D boxes into the detected 2D
boxes. A similar approach introduced by \cite{DBLP:conf/cvpr/MousavianAFK17} is
to extend a 2D detector to also estimate the dimensions and the orientation of a
vehicle. In a second step, a 3D box with the estimated parameters is fitted
into the 2D box. \cite{DBLP:conf/eccv/ZhouKK20} solved the problem of 3D object
detection by estimating the center point of an object and regressing its
parameters. Another approach, proposed by \cite{Brazil2019ICCV}, is to use a
region proposal network as introduced by
Faster~R-CNN~\cite{DBLP:conf/nips/RenHGS15} to estimate 3D parameters. These
approaches rely on the availability of copious training data to be viable. As
mentioned before, real-world training data which is annotated with 3D data is very
scarce.

One approach to solve the association problem is to compare the appearance of
pairs of detected objects. For example, \cite{DBLP:conf/cvpr/Leal-TaixeCS16}
train a Siamese CNN which aims to associate pedestrian detections. However, in
our scenario we track objects where occlusion is a major issue; in heavy
traffic, vehicles frequently occlude each other. Furthermore, maneuvering
vehicles can quickly change their appearance. We therefore refrain from
considering appearance features for association purposes altogether and instead
adopt a pipeline similar to SORT~\cite{DBLP:conf/icip/BewleyGORU16}. The key
idea is to use the bounding boxes predicted by the Kalman filter for
association purposes. Unlike \cite{DBLP:conf/icip/BewleyGORU16}, we use a motion
model in 3D space instead of 2D space.

The possibility of modeling maneuvering vehicles with a coordinated turn model
has been used in the context of
aircrafts~\cite{gertz1989multisensor,DBLP:conf/fusion/RothHG14} and ground
vehicles~\cite{kim2004imm}. These systems typically employ a non-linear variant
of the Kalman filter~\cite{kalman1960new} to track vehicles. However,
while most of these approaches rely on radar sensors or lidar sensors, our
approach requires only a video camera. It uses sequences of 2D detections and the
camera calibration in order to approximate 3D states.

Several traffic scenarios similar to ours have been studied. One commonly used
benchmark is the tracking benchmark of the KITTI dataset~\cite{Geiger2012CVPR}.
However, this benchmark evaluates tracks of 2D boxes only, while we are
interested in tracks of 3D boxes. Furthermore, the video sequences were recorded
from a camera located inside the car, which differs from the typically elevated
perspective of our scenario. Another tracking benchmark is part of the UA-DETRAC
dataset~\cite{DBLP:journals/cviu/WenDCLCQLYL20}. Similarly to our scenario, this
dataset consists of video sequences recorded by stationary traffic surveillance
camera. However, since there is no ground truth data for 3D boxes, we cannot use
this benchmark to evaluate our 3D tracking performance either. In addition to
that, the camera calibration is unknown.

%% file: sourcefiles/approach.tex
\section{Approach}

\emph{Urban Traffic Surveillance} (UTS) relies solely on a calibrated monocular video camera as the main sensor. The intrinsic calibration is given by the camera matrix $\mathbf{K} \in \mathbb{R}^{3 \times 3}$ that describes the projection of a pinhole camera from a \emph{camera coordinate system} (CCS) to an \emph{image coordinate system} (ICS):
\begin{align}
	\begin{pmatrix}
		\vec{x}_\tinytext{ics} \\
		1
	\end{pmatrix}
	\propto \mathbf{K} \cdot \vec{x}_\tinytext{ccs}.
\end{align}

The extrinsic calibration is given by the projection matrix $\mathbf{P} \in \mathbb{R}^{3 \times 4}$ which contains the orientation and offset of the CCS to a camera independent \emph{world coordinate system} (WCS):
\begin{align}
	\vec{x}_\tinytext{wcs} \definedas \mathbf{P} \cdot
	\begin{pmatrix}
		\vec{x}_\tinytext{ccs} \\
		1
	\end{pmatrix}.
\end{align}
The $z$-axis of the WCS is perpendicular to the street plane. Thus, the $x$- and $y$-axis span the street plane which we assume to be perfectly planar.

An additional object specific \emph{vehicle coordinate system} (VCS) is defined w.r.t. the WCS. The vehicle center $\vec{c}_{\tinytext{wcs}}$ is the origin and its orientation $\phi$ inside the street plane defines the relative rotation matrix $\mathbf{R}_{z}(\phi)$ that rotates around the normal vector of the street plane:
\begin{align}
	\vec{x}_{\tinytext{wcs}} \definedas \mathbf{R}_{z}(\phi) \cdot \vec{x}_{\tinytext{vcs}} + \vec{c}_{\tinytext{wcs}}.
\end{align}

\subsection{Detection, Classification \& Prior Knowledge}
\label{sec:detection-classification-and-prior-knowledge}

The detector is the only component to process the raw data of the video camera and by that exploit its full information. We base our approach on the YOLOv3~\cite{redmon2018CoRR} framework which is a one shot deep network that employs an anchor related regression to perform detection and classification. The detector abstracts the information input of our surveillance system from a visual representation to a purely geometric one by identifying all objects of interest and replacing their visual representations by two-dimensional boxes, classification labels and scores. The two-dimensional and axis-aligned bounding boxes (i.e. the measurements) in the image domain are given by the pixel coordinates of the four edges \emph{top}, \emph{left}, \emph{bottom} and \emph{right}:
\begin{align}
	 \vec{m} \definedas \left( t, l, b, r \right)^{T}.
\end{align}
This abstraction to a purely geometric approach enables a real-time application even with limited hardware resources.

The objects of interest for our application are only vehicles. Thus, we dismiss all detections that are not classified as \texttt{CAR}, \texttt{TRUCK} or \texttt{BUS} by the detector. For these remaining vehicle classes, we introduce class specific prior knowledge about the shape $\vec{s}$ which consists of the values \emph{length}, \emph{width} and \emph{height}. The prior knowledge is given by the mean $\hat{\vec{s}}$ and a covariance matrix $\hat{\mathbf{C}}_{\vec{s}} \equiv \mathbf{C}_{\tinytext{prior}}$. The magnitude of each of the class specific covariance matrices reflects the specificity of the shape within the corresponding class.

\subsection{State, Observation \& Motion Model}
\label{sec:state-observation-and-motion-model}

The observation model relates the internal vehicle state representation $\vec{x}$ from state space to the detection $\vec{m}$ given in the observable measurement space. The motion model is used to propagate the internal state over time.

For the state and the observation and motion model, we need to differentiate between two stages of an observed object.

\subsubsection{Initial 2D Stage}

From the moment of detection until a \emph{suitable amount of motion} is achieved, the representation of the vehicle is only given by an axis-aligned 2D bounding box that may perform a two-dimensional accelerated translation in the image domain and may be affected by a change of scale. During this stage, the state and its motion model is given by
\begin{align}
	\vec{x}_{\tinytext{2D},t} \definedas
	\begin{pmatrix}
		\vec{c} \\
		\vec{s} \\
		\vec{v} \\
		\Delta \vec{s} \\
		\vec{a}
	\end{pmatrix}_{t}
	\stackrel{f_{\tau}}{\rightarrow}
	\begin{pmatrix}
		\vec{c} + \vec{v} \cdot \tau + \frac{1}{2} \cdot \vec{a} \cdot \tau^{2} \\
		\vec{s} \odot \exp(\Delta \vec{s} \cdot \tau) \\
		\vec{v} + \vec{a} \cdot \tau \\
		\Delta \vec{s} \\
		\vec{a}
	\end{pmatrix}_{t + \tau}
	\definedas \vec{x}_{\tinytext{2D}, t + \tau},
\end{align}
where $f_{\tau}$ is the state transition function that represents the motion model and propagates a given state by a time fraction $\tau$. $\odot$ denotes elementwise multiplication, $\vec{c}$ consists of the 2D pixel coordinates of the box center, $\vec{s}$ is the pixel width and height of the box, $\vec{v}$ and $\vec{a}$ are the 2D pixel velocity and acceleration of the center and $\Delta \vec{s}$ contains the logarithmic rate of the scale change for width and height respectively. The 2D observation model is just given by the mapping $h$
\begin{align}
	\vec{x}_{\tinytext{2D}} \stackrel{h}{\rightarrow} \left( \vec{c}^{T}, \vec{s}^{T} \right)^{T} = \vec{y},
\end{align}
which gives the representation $\vec{y}$ in measurement space for a state $\vec{x}$ in the state space. The mapping extracts the position and shape components from the state vector that is already defined in the ICS as is the detection.

\subsubsection{3D Stage}

Once a vehicle was tracked several times to the point that an estimation of the direction of movement in the street plane yields an uncertainty below a threshold, we replace the basic two-dimensional representation by a more expressive three-dimensional one. This allows us to introduce a more sophisticated motion model which describes the motion in the space where it is physically explainable by a model of an inert vehicle with a given momentum and steering angle. The state and its transition function $f_{\tau}$ are given by
\begin{align}
	\vec{x}_{\tinytext{3D},t} \definedas
	\begin{pmatrix}
		\vec{c} \\
		\vec{s} \\
		\phi \\
		v \\
		\omega
	\end{pmatrix}_{t}
	\stackrel{f_{\tau}}{\rightarrow}
	\begin{pmatrix}
		\vec{c} + \frac{v}{\omega}
		\begin{pmatrix}
			\sin(\phi + \omega \tau) - \sin(\phi) \\
			\cos(\phi) - \cos(\phi + \omega \tau)
		\end{pmatrix}
		\\
		\vec{s} \\
		\phi + \omega \tau \\
		v \\
		\omega
	\end{pmatrix}_{t + \tau}
	\definedas \vec{x}_{\tinytext{3D}, t + \tau},
\end{align}
where $\vec{c}$ is the vehicle bounding box center in the WCS, $\vec{s}$ is the shape (length, width, height), $\phi$ is the orientation within the street plane, $v$ the velocity and $\omega$ the angular velocity. This motion model is called the \emph{coordinated turn model} with polar velocity as proposed in \cite{gertz1989multisensor}. It describes an orientated object that incrementally moves with a constant turning rate $\omega$ and velocity $v$ on a circular arc.

The 3D representation allows a much more meaningful depiction of the motion but comes at the cost of a more complex observation model compared to the 2D stage. The highly nonlinear observation model is given by the mapping $h$
\begin{align}
	\vec{x}_{\tinytext{3D}}	\stackrel{h}{\rightarrow} \left( t, l, b, r \right)^{T} = \vec{y},
\end{align}
which projects the eight edges of the 3D bounding box from the VCS to the ICS and identifies the horizontal and vertical minimum and maximum components that define the extent of the framing 2D bounding box.

Both models explicitly consider the time $\tau$ that has elapsed. This allows our system to easily deal with unstable frame rates, frame drops or missing detections.

\subsection{Association}
\label{sec:association}

For every new video frame, the detector component generates a set of new detections $\{ \vec{m}_{t + \tau, i} \}_{i}$ (classified 2D bounding boxes). The task of the association component is to assign those new detections to the existing vehicle observations given by their states $\{ \vec{x}_{t, j} \}_{j}$, or create a new one. According to whether the state is already given as a 3D or as an early 2D representation (see \ref{sec:state-observation-and-motion-model}), the respective state transition function $f_{\tau}$ and observation model $h$ are used to generate prediction $\{ \vec{y}_{t + \tau, j} \}_{j}$ of the vehicle states in the measurement space with
\begin{align}
	\label{eq:measurement-prediction}
	\vec{y}_{t + \tau}= h\left(f_{\tau}\left(\vec{x}_{t}\right)\right).
\end{align}

We use the \emph{Hungarian algorithm}~\cite{kuhn1955hungarian} to determine a set of matches $\vec{m}_{i} \leftrightarrow \vec{y}_{j}$ so that the total costs of assignment are minimum. The cost of a potential assignment is based on the \emph{intersection over union} (IoU) of the respective 2D bounding boxes.

\subsection{3D-Initialization}
\label{sec:3d-initialization}

Every vehicle only consists of a 2D representation as long as not enough motion has been accumulated during the observation in the ICS. Once this is the case, we utilize the class specific shape prior $\hat{\vec{s}}$, $\mathbf{C}_{\tinytext{prior}}$ of the majority classification result of the detections of this vehicle to get an estimate of the orientation $\phi$ of the vehicle.

\subsubsection{Orientation Estimation}

The two detections $\vec{m}_{t}$ and $\vec{m}_{t + \tau}$ which differ by significantly more translation in the ICS than is explainable by detection noise are used to get an initial estimate for the orientation angle $\phi$. We assume that the two 2D bounding box centers $\vec{c}_{\tinytext{ics}, t}$ and $\vec{c}_{\tinytext{ics}, t + \tau}$ are also suitable estimates for the projection of the 3D bounding box centers $\vec{c}_{\tinytext{wcs}, t}$ and $\vec{c}_{\tinytext{wcs}, t + \tau}$. The 3D reconstruction exploits the shape prior to compute the world coordinates such that the centers are located in a height of $\hat{h}/2$ above street level. These reconstructions define the initial orientation estimate $\phi$ and also the velocity estimate $v$ of the vehicle under the assumption that direction of travel and longitudinal axis coincide.

\subsubsection{Active Corner Estimation}
\label{sec:active-corner-estimation}

\begin{figure}[tb!]
	\centering
	\includegraphics[width = 0.49\linewidth]{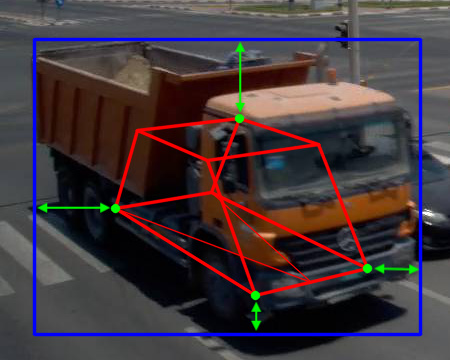}
	\includegraphics[width = 0.49\linewidth]{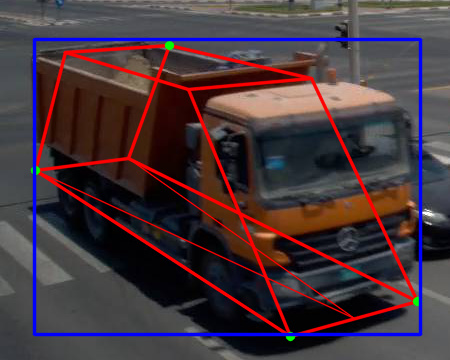}
	\caption{The left images shows the prior 3D bounding box (red) inside the 2D detection box (blue). The active corners which define the outline of the 3D box are highlighted in green. The right image depicts the 3D box after its shape and center have been optimized to minimize the deviation to the detection box. Best viewed in color.}
	\label{fig:init-with-prior-and-les}
\end{figure}

We now have an initial estimate of the two oriented 3D bounding boxes that are related to the two measurements $\vec{m}_{t}$ and $\vec{m}_{t + \tau}$. They share the center position with their corresponding detection but the extent is only given by the class specific but still very coarse prior shape $\hat{\vec{s}}$ that might not be in good agreement with the detection (see Fig.~\ref{fig:init-with-prior-and-les}). To further improve this coarse estimate, the main idea of the 3D initialization is to vary the center $\vec{c}_{\tinytext{wcs}, t}$, the displacement given by the velocity $v$ and of course the shape $\vec{s} = (l, w, h)^{T}$ such that the resulting 3D boxes fit best into the 2D detection boxes. The orientation is held fixed to avoid getting a nonlinear problem.

We use the initial 3D bounding boxes to identify the subset of its eight vertices that define the outline that we wish to match with the detection (see Fig.~\ref{fig:init-with-prior-and-les}).

\subsubsection{Least Squares Optimization}
\label{sec:least-squares-optimization}

The problem formulation consists of six unknowns ($x$-$y$ center position, shape, velocity) and eight equations with each four of them defined by the respective detection. The six unknowns and the fixed orientation define the location of the \emph{active corners} (see \ref{sec:active-corner-estimation}) which combined with the intrinsic and extrinsic calibration define the resulting outline. Rearranging the given information and the unknowns yields an overdetermined linear equation system that can be directly solved as the least squares minimization problem
\begin{align}
	\label{eq:least-squares-optimizaiton}
	l_{\tinytext{detection}} \definedas \left\| \mathbf{M} \cdot
	\begin{pmatrix}
		\vec{c}_{\tinytext{wcs}, t} \\
		\vec{s} \\
		v
	\end{pmatrix}
	- \vec{b} \right\| \rightarrow \min,
\end{align}
where the matrix $\mathbf{M} \in \mathbb{R}^{8 \times 6}$ and the vector $\vec{b} \in \mathbb{R}^{8 \times 1}$ are completely defined by the known quantities $\mathbf{P}$, $\mathbf{K}$, $\phi$, $\vec{m}_{t}$ and $\vec{m}_{t + \tau}$.

\subsubsection{L2 Regularization with Shape Prior}

\begin{figure}[tb!]
	\centering
	\includegraphics[width = 0.49\linewidth]{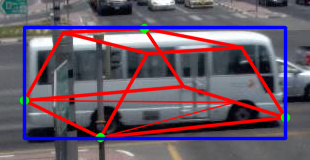}
	\includegraphics[width = 0.49\linewidth]{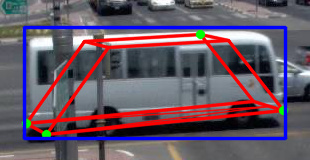}
	\caption{On the left-hand side, one can see the resulting 3D box (red) if the only objective is to fit best into the detection box (blue). On the right-hand side, the minimization has been regularized to restrict the shape utilizing a class specific prior. Best viewed in color.}
	\label{fig:init-with-les-regularized}
\end{figure}

The resulting 3D bounding box given as the solution of \eqref{eq:least-squares-optimizaiton} is just defined by the task to achieve the best match with the detection box. So there is no restriction on the dimension or the aspect ratio of the shape. Since the detection is affected by noise and since we also have prior knowledge about the shape of a vehicle of a specific class, we include a regularization term to the loss function \eqref{eq:least-squares-optimizaiton} and weigh each term with a covariance matrix that represents the uncertainty of the respective objective:
\begin{align}
	\label{eq:least-squares-with-regularization}
	\begin{split}
		l_{\tinytext{detection + prior}} \definedas& \left\| \mathbf{M} \cdot
		\begin{pmatrix}
			\vec{c}_{\tinytext{wcs}, t} \\
			\vec{s} \\
			v
		\end{pmatrix}
		- \vec{b} \right\|^{2}_{\mathbf{C}_{\tinytext{detection}}} + \\
		\phantom{\definedas}& \left\| \vec{s} - \hat{\vec{s}} \right\|^{2}_{\mathbf{C}_{\tinytext{prior}}} \rightarrow \min
		\end{split}.
\end{align}
The solution given by the minimum of \eqref{eq:least-squares-with-regularization} is a 3D bounding box that is not only defined by the noisy detection but it is in best agreement with the two objectives of matching the detection box and fulfilling the statistical prior knowledge about the vehicle shape (compare both cases in Fig.~\ref{fig:init-with-les-regularized}).

\subsection{Kalman Filter Update}
\label{sec:kalman-filter-update}

Whenever a new video frame is being processed, the existing vehicle states $\vec{x}_{t}$ and their uncertainties $\mathbf{C}_{\vec{x}_{t}}$ are propagated from $t$ to $t + \tau$ using the state transition function $f_{\tau}$ and a suitable error propagation method (either directly, by linearization or by sampling). If the association component (see section \ref{sec:association}) assigns a new detection $\vec{m}_{t + \tau}$ to an existing vehicle a Kalman filter is used to perform the data fusion of the detection given in the measurement space and the propagated vehicle state $\hat{\vec{x}}_{t + \tau}$ given in state space.

In case the vehicle is still in the early 2D stage (see section \ref{sec:state-observation-and-motion-model}), its observation model is strictly linear and the motion model can be approximated sufficiently by a linearization. The extended Kalman filter (EKF) is used to compute the fusion and yield the updated state estimate $\vec{x}_{t + \tau}$ and its uncertainty $\mathbf{C}_{\vec{x}_{t + \tau}}$.

Otherwise, if the vehicle already reached the more advanced 3D stage, the limitations of an EKF become more apparent with both the observation and motion model being strongly non-linear. Especially when the vehicle orientation is massively changing (e.g. during a U-turn or sharp turning maneuver) the EKF fails to deliver a good update. This is why we use an unscented Kalman filter (UKF)~\cite{julier1997new} during the 3D stage. It performs the propagation without linearization but by a sampling of carefully selected sigma points which are then mapped by the state transition function and finally mean and covariance of this discretized distribution are re-estimated. The UKF is capable of updating the internal states with new detections in good agreement with the motion model even if the vehicle is performing strong turning maneuvers.

\subsection{Miscellaneous / Challenges}
\label{sec:miscellaneous}

\subsubsection{Occlusions}

\begin{figure}[tb!]
	\centering
	\includegraphics[width = \linewidth]{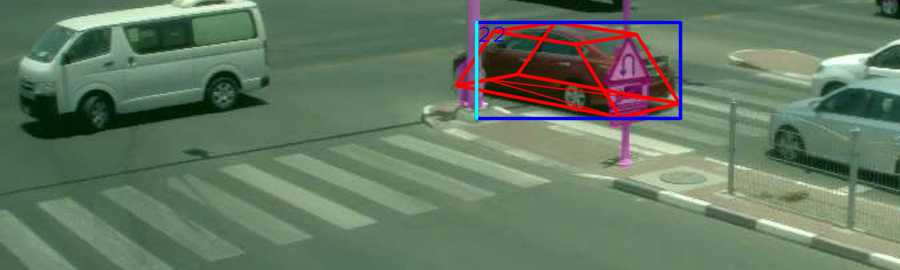}
	\\
	\vspace{0.1cm}
	\includegraphics[width = \linewidth]{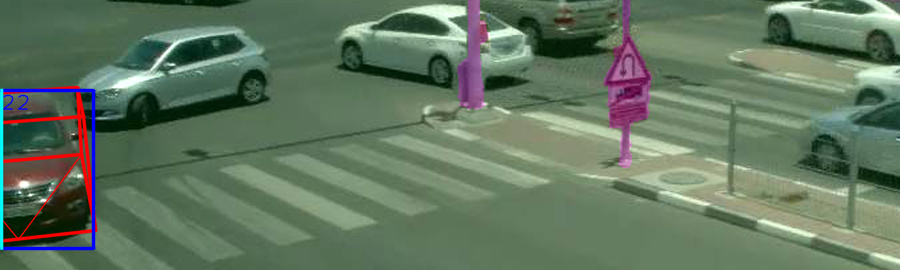}
	\caption{The red car is affected by two types of occlusion while performing a U-turn maneuver. The bottom image depicts the vehicle being partially outside the image region. The affected edge (highlighted in cyan) that violates an image border threshold is excluded and not used by succeeding components. The top image visualizes occlusion by static infrastructure (area colored in magenta). The affected edge (highlighted in cyan) overlaps by more than a threshold with the potentially occluding area and is dismissed. Best viewed in color.}
	\label{fig:occlusion}
\end{figure}
Apart from noise, the detection box may also be strongly affected by occlusion. In contrast to noise which we model as a zero-mean Gaussian with a few pixels of standard deviation, occlusions will have a more severe impact on the measurements and cannot be easily described by white noise. We observe three different cases when the detection box does not completely cover the vehicle such that a subset of the edges does not define the outline of the vehicle:
\begin{itemize}
	\item The first case is the easiest to detect and actually is not an occlusion as such. When the vehicle partially leaves or enters the image region it cannot be completely observed by the detector. We simply dismiss all edges with an image border distance below a threshold and exclude them from later usage (see bottom image in Fig.~\ref{fig:occlusion}).
	\item The second case happens when the vehicle is partly blocked by static infrastructure (e.g. traffic lights, walls, buildings). Since our surveillance cameras are fixed and always observe the same area, we semantically labeled all image regions that may cause occlusion (see masked areas in magenta in Fig.~\ref{fig:occlusion}). We dismiss those edges of a detection that possess an overlap of more than 50\% with potentially occluding areas (see top image in Fig.~\ref{fig:occlusion}). This procedure is performed solely in the ICS which has the advantage that the edges of the detection can be validated within the detector component. On the other hand, by that we do not exploit 3D information (i.e. the object distance to the camera) and cannot verify if the vehicle is occluded by infrastructure or the other way round. Nevertheless, we might lose some incorrectly classified edges of the detection but static occlusion is effectively prevented.
	\item The last case we observed is dynamic occlusion which we do not take care of at the moment. In case the occlusion is caused by other vehicles, the estimated 3D state information can be used to identify which object is blocking the other and which edges are affected and to be dismissed. 
\end{itemize}
During the 3D initialization in section \ref{sec:3d-initialization}, we only rely on detections with all four edges being valid and not affected by occlusion. Later, when the vehicle state is tracked and updated by utilizing the Kalman filter (see section \ref{sec:kalman-filter-update}), we simply modify the observation model $h$, such that only the valid edges of the detection / measurement are recognized.

\subsubsection{Error Propagation}

At all times, the input and output of every component which is affected by noise is described by a mean and a covariance matrix. This yields a fully probabilistic approach that always allows to judge the quality of an estimate and depending on that decide how to proceed. Noise is only defined when it initially takes influence in the process. For example, the detection is given by the four pixel coordinates of the edges (see section \ref{sec:detection-classification-and-prior-knowledge}). We assume that all values are overlaid with white Gaussian noise. When first estimating the vehicle orientation and velocity in section \ref{sec:3d-initialization}, we propagate the detector noise (and the uncertainty of the height prior) to the output.

Whenever possible and efficient, this is done directly by deriving the covariance of the output from the covariance of the input. Otherwise, the propagation is achieved by linearization and propagation of uncertainty using the Jacobians of the mapping. In case of a strongly nonlinear relation, a sampling which exploits the known statistics of the input is performed and the covariance is estimated from the projected output samples.

%% file: sourcefiles/experiments.tex
\section{Experiments}

As mentioned earlier, the difficulty of obtaining annotated 3D data of a traffic
surveillance scenario is a major obstacle to the quantitative evaluation of
systems like ours. Instead of investing considerable resources into the
annotation of real-world data, we opted for the more pragmatic approach of
generating a synthetic dataset and evaluating our system on that. We believe
that the challenge of performing well on a synthetic dataset should in many
cases also translate to a good performance on real-world data.

For this reason, we have used the CARLA platform \cite{Dosovitskiy2017PMLR} for
our purposes. The intention of CARLA is to provide a platform on which one can
develop autonomous driving agents. It includes a simple driving logic that can
be used to control virtual vehicles which follow random paths around a virtual
city, while at the same time obeying the traffic laws and not colliding with
each other. Furthermore, it comes with a set of virtual cities and vehicle
models. This allows us to create a completely virtual environment to evaluate
our system. We have placed virtual cameras at spots around a road junction which
were supposed to roughly reflect the spots on which one would install traffic
surveillance cameras in the real world.

Besides making the generation of arbitrary synthetic data a trivial task, our
approach also has the advantage that we are operating in a controlled
environment. That means we can eliminate other error sources such as calibration
errors of the camera and we can model the roads as perfectly flat planes such
that the ground plane assumption holds. By doing that, we can iteratively
improve our system such that it first performs well when running under perfect
conditions and subsequently make it more robust to also handle realistic
conditions. In a real-world scenario, we have to take into account that the
calibration of a camera always contains a certain degree of error and that roads
do not tend to be perfectly flat and even.

In order to evaluate the tracking performance of our system, we have modified
the evaluation script of the KITTI object tracking benchmark
\cite{Geiger2012CVPR}. This benchmark is a pure 2D tracking benchmark. During
evaluation, the tracked trajectories of 2D boxes are matched to ground truth
trajectories. This matching occurs by considering the intersection over union (IoU)
between pairs of tracked 2D boxes and ground truth 2D boxes. On the basis of
this matching, evaluation metrics are then computed. Inspired by
\cite{Weng2020_AB3DMOT_eccvw}, the fundamental change we performed to evaluate
the performance of 3D tracking was a pairwise comparison of 3D boxes instead of 2D boxes
using a three-dimensional IoU.

\begin{figure}[t!]
  \centering
  \includegraphics[width=\linewidth]{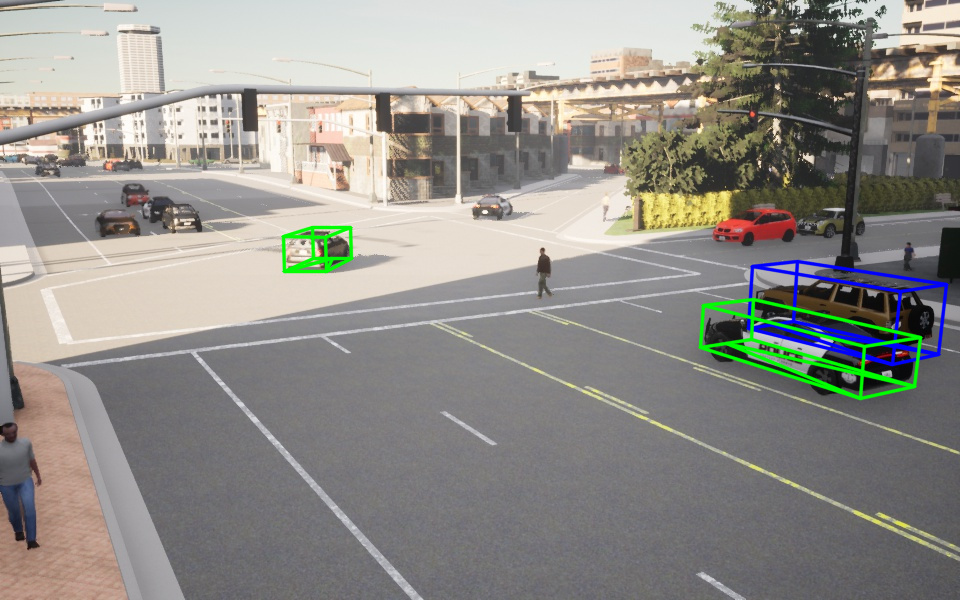}\\[2mm]
  \includegraphics[width=\linewidth]{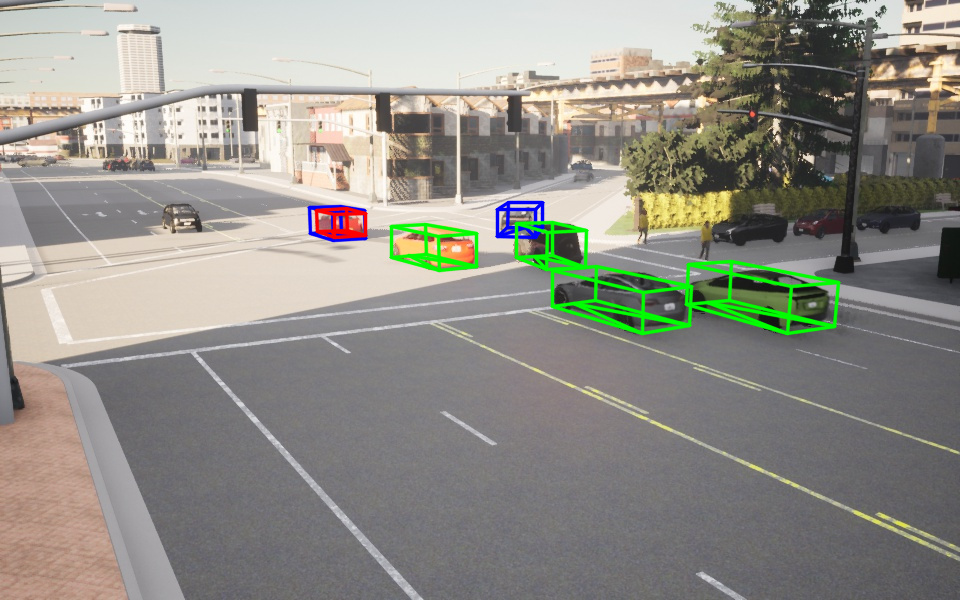}
  \caption{These two frames exemplary visualize the evaluation process. Green boxes
    represent successful detections, blue boxes are false negatives and red
    boxes are false positives. Best viewed in color.}
  \label{fig:evaluation}
\end{figure}

We have generated a two minute sequence of a $960 \times 600$ video running with
20 frames per second. Our implementation of UTS is capable of processing
approximately 20 frames per second on a workstation with an Intel Core i7-6700k
CPU and a NVIDIA GTX 1050 TI GPU. Two frames of our dataset are depicted in
Fig.~\ref{fig:evaluation}. These specific frames were purposely chosen to
highlight two of the major difficulties of the benchmark:
\begin{itemize}
  \item Vehicles being occluded by other vehicles or by fixed objects. This
    problem is not so easy to deal with using only a single camera. Perhaps it
    could be tackled by using a detector specifically tailored for this scenario
    and not a general detector such as the YOLOv3 detector that we employed.
  \item Since surveillance cameras often tend to be installed in a way such that
    the optical axis intersects the road plane in a acute angle, small errors in
    the 2D detection can induce large errors in world coordinates. This problem
    gets even more significant in a real-world scenario when other uncertainties
    such as calibration errors and uneven roads are introduced.
\end{itemize}

The evaluations metrics output by the KITTI tracking benchmark script are
depicted in Table~\ref{tbl:evaluation}. The MOTA metric
\cite{DBLP:journals/ejivp/BernardinS08} counts the number of false positives,
false negatives and ID switches and has a maximum value of 1. The \emph{Mostly Tracked}
and \emph{Mostly Lost} metrics \cite{DBLP:conf/cvpr/LiHN09} indicate the fraction of
ground truth tracks which have been tracked for more than 80\% or by less than
20\% of their whole trajectories, respectively.

As we have mentioned above, the benchmark first matches bounding boxes found by
the tracker with ground truth bounding boxes and subsequently calculates
evaluation metrics. In order for a match to be accepted, the IoU of the
corresponding boxes have to exceed a certain threshold. This threshold is set to
$0.5$ in the KITTI tracking benchmark. However, \cite{Weng2020_AB3DMOT_eccvw}
chose the threshold $0.25$ for their 3D tracking benchmark. As we can see in
Table~\ref{tbl:evaluation}, it is indeed sensible to choose a threshold lower
than $0.5$ when matching 3D boxes, since the task of accurately tracking 3D
boxes is significantly more difficult and the IoU metric is more sensitive to deviations in higher dimensional cases.
\begin{table}[b!]
	\begin{tabular}{r || r | r | r | r}
		IOU & & Mostly & Partly & Mostly \\
		thresh. & MOTA & Tracked & Tracked & Lost \\\hline
		0.5 & 0.158823 & 0.279070 & 0.441860 & 0.279070 \\
		0.25 & 0.617005 & 0.558140 & 0.348837 & 0.093023 \\
		0.1 & 0.693864 & 0.581395 & 0.348837 & 0.069767
	\end{tabular}
	\caption{Evaluation result}
	\label{tbl:evaluation}
\end{table}

It is difficult to compare these results with other approaches, because as we
have stated earlier, to the best of our knowledge no other public benchmark with
this scenario (stationary traffic camera with 3D ground truth data) exists.
\cite{Weng2020_AB3DMOT_eccvw} uses the KITTI dataset, which was created by
cameras mounted on moving vehicles. However, it is still possible to get an
idea about the performance of UTS by considering the results in
Table~\ref{tbl:evaluation}. As we have argued above, it is reasonable to choose
$0.25$ as IoU threshold for a successful matching. Then, we can see that
approximately 56\% of all tracks were tracked successfully, 35\% were tracked
partly and only 9\% have not been tracked correctly for the most part. Upon further observation, we found that in
several cases a vehicle appeared in the ground truth data but was not detected
by our system simply because it was located at the edge of the detection area
that we consider (green area in Fig.~\ref{fig:teaser}). All vehicles that our detector located just slightly outside
of this area are discarded. For a fair evaluation, it might make sense to take
the same approach as KITTI and define ``Don't care''-areas at the edge of the
detection area such that a tracker will not get penalized in cases where it
discards vehicles which are just slightly inside the detection area. We believe
that incorporating these ``Don't care''-areas in our benchmark in the future
will result in fairer comparison and better results, because then a tracker gets punished only in the
case where it loses its target while it is located inside the part of the
junction which we consider. That change should result in several of the
trajectories that are currently considered to be partly tracked to become mostly
tracked instead.

\begin{figure}[t!]
	\centering
	\includegraphics[width = 0.49\linewidth]{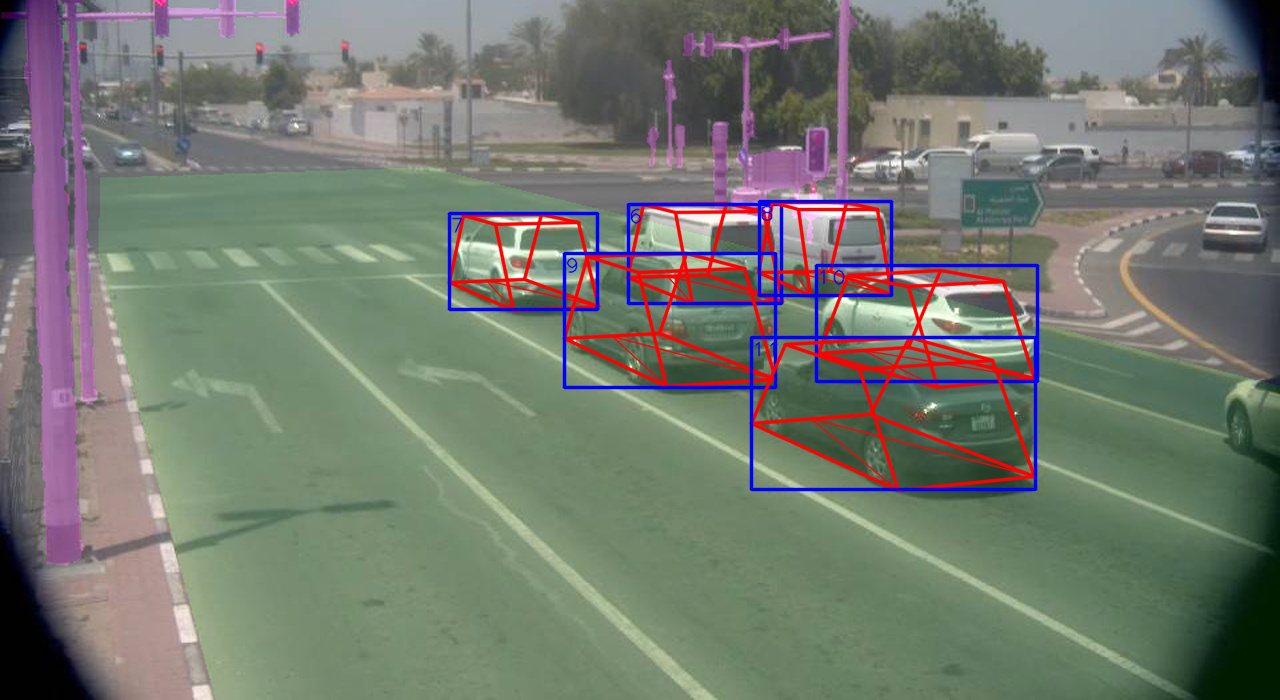}
	\includegraphics[width = 0.49\linewidth]{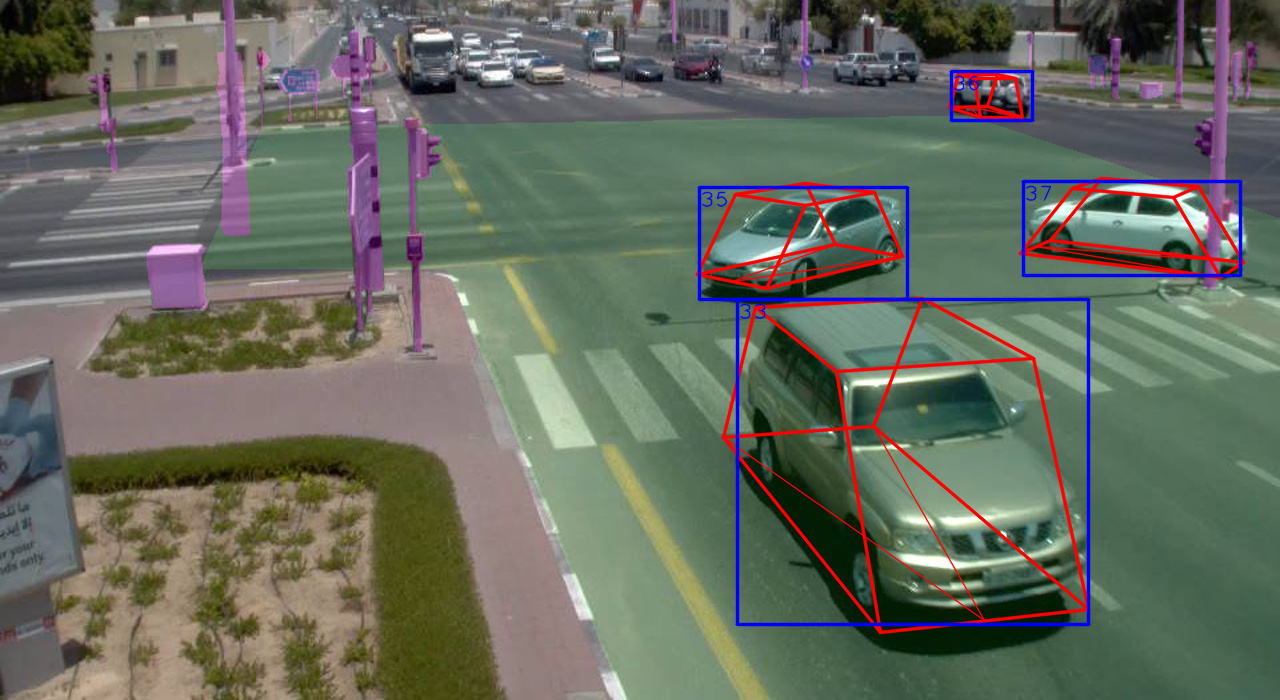}
	\\
	\vspace{0.1cm}
	\includegraphics[width = 0.49\linewidth]{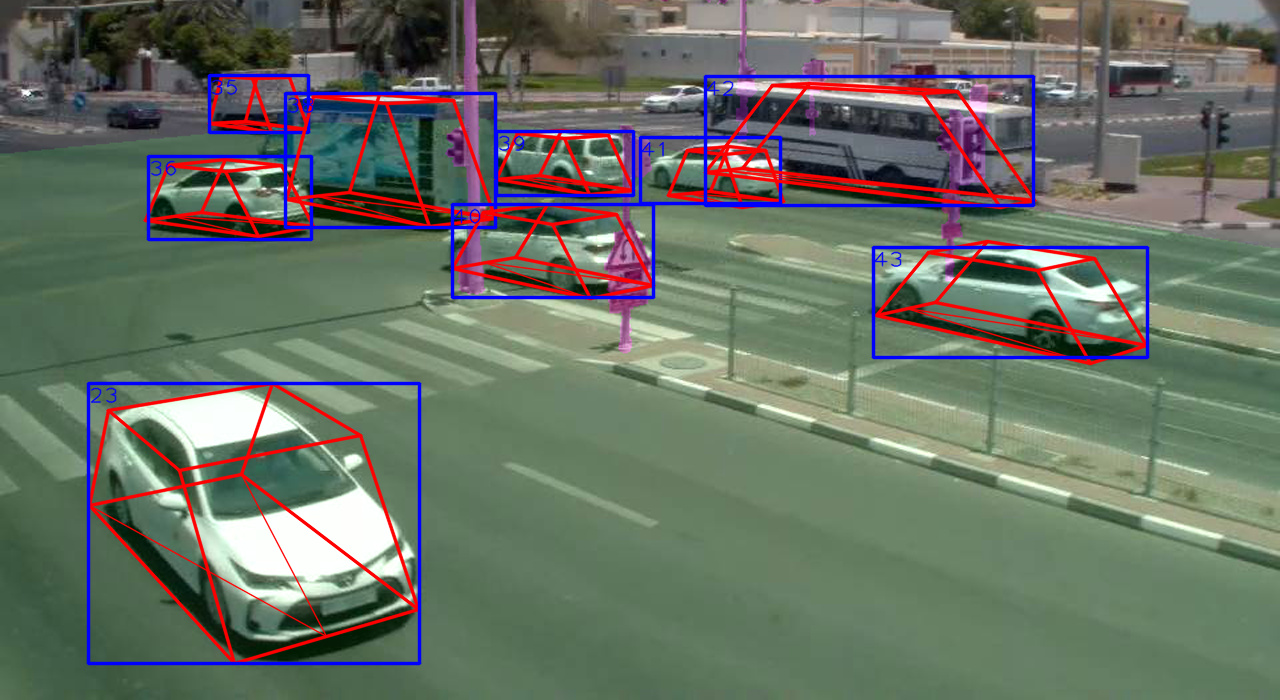}
	\includegraphics[width = 0.49\linewidth]{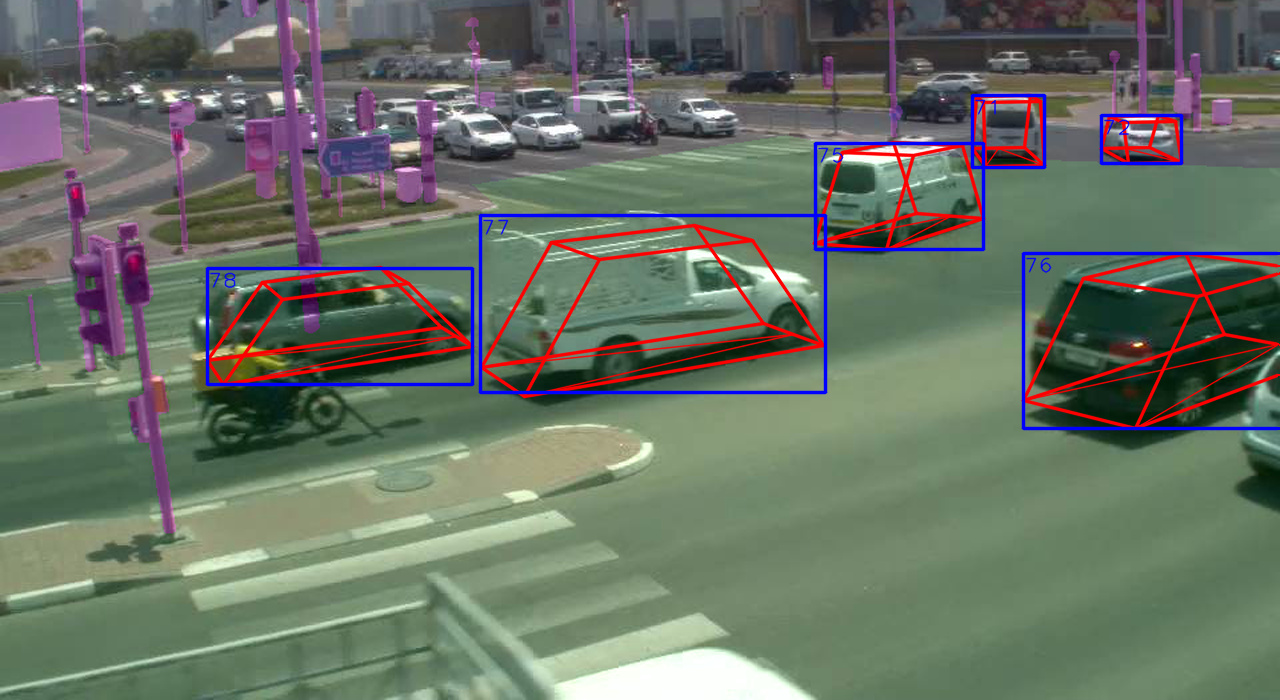}
	\\
	\vspace{0.1cm}
	\includegraphics[width = 0.49\linewidth]{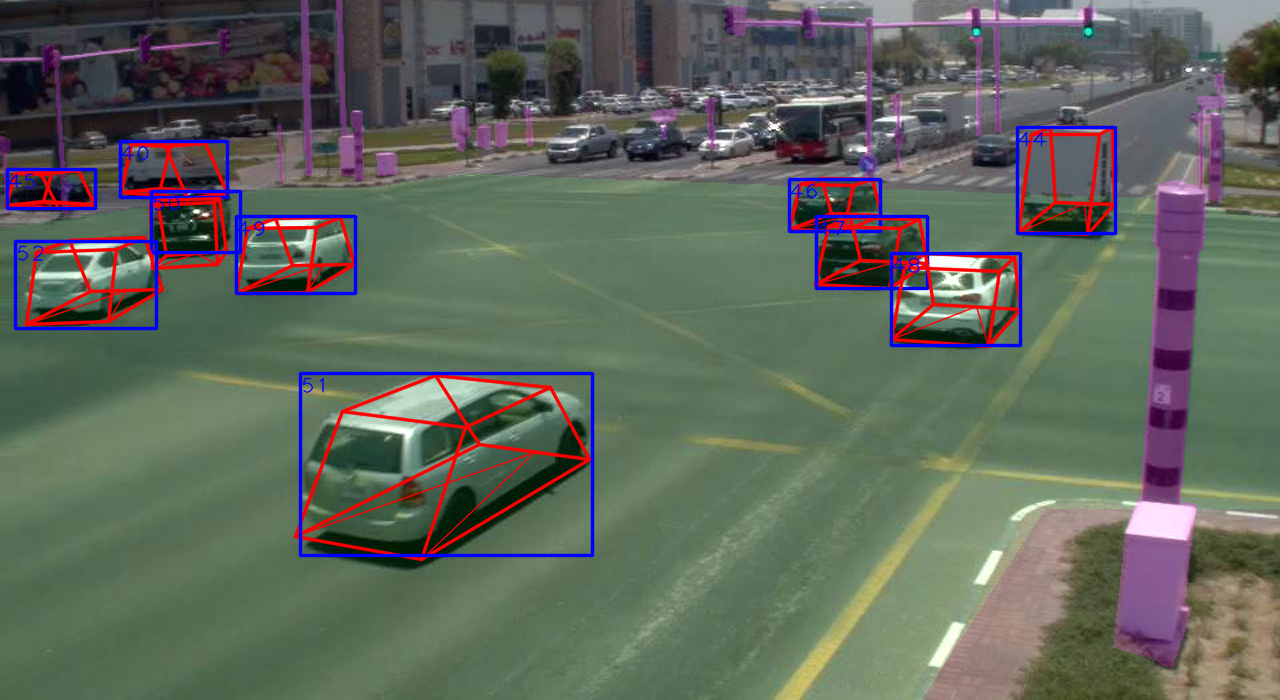}
	\includegraphics[width = 0.49\linewidth]{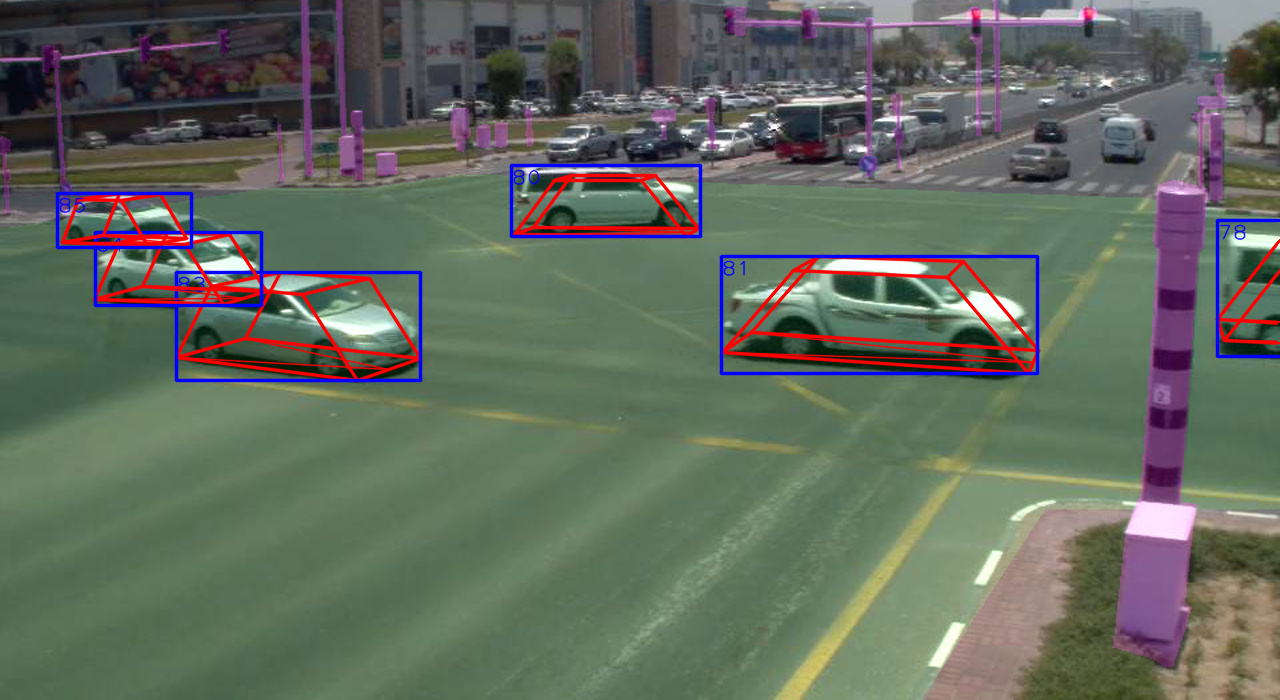}
	\caption{These images serve to give an impression of how UTS performs on real-world data. The detection boxes are depicted in blue and the tracked 3D bounding boxes are drawn in red color. The area masked in light green color defines the area inside of which our system actively performs detection and tracking. Areas semantically labeled in magenta may potentially occlude vehicles passing by. Best viewed in color.}
	\label{fig:experiments-qualitative-general-case}
\end{figure}
Apart from the advantages of synthetic data to perform quantitative evaluation and software development under controlled conditions, UTS was designed with the goal of a real-world application executed on real-world data. For that purpose, the performance of UTS was qualitatively evaluated on real-world data in parallel during the development process of the surveillance system. By this, we ensured that UTS was designed to yield a robust performance even on more demanding datasets where the ground plane assumption is not perfectly fulfilled and the camera calibration contains a certain amount of noise. The real-world dataset also consists of a similar urban road intersection with cameras mounted in similar elevated positions all over the scene. Fig.~\ref{fig:experiments-qualitative-general-case} gives a good impression of the general detection and tracking performance. Also in this real-world scenario UTS achieves a consistent 3D tracking performance. We can deduce this from the fact that the tracked 3D bounding boxes are consistent over time and coherent with the 2D detection. This is only possible if the detections in the image domain, the ground plane given by the intrinsic and extrinsic calibration, and the motion model defined in the world coordinate system are in good agreement and compliant with the prior knowledge about the vehicle shape.

\begin{figure}[t!]
	\centering
	\includegraphics[width = 0.32\linewidth]{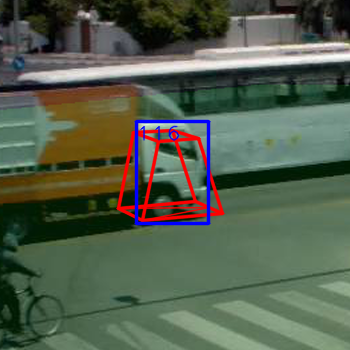}
	\includegraphics[width = 0.32\linewidth]{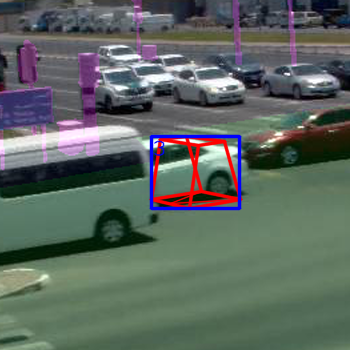}
	\includegraphics[width = 0.32\linewidth]{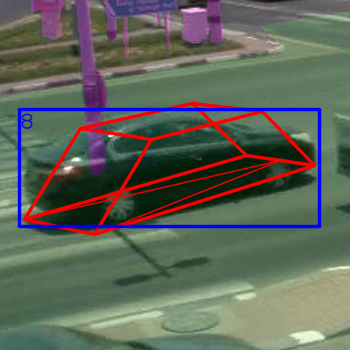}
	\caption{The dominant cases when UTS yields erroneous 3D tracking results are detection errors especially with large vehicles as trucks and buses (left), dynamic occlusions caused by other vehicles (middle) or orientation initialization errors (right).}
	\label{fig:experiments-qualitative-error-case}
\end{figure}

The exemplarily chosen images in Fig.~\ref{fig:experiments-qualitative-general-case} do not represent a cherry picking of a rarely occurring best-case scenario but they represent the general performance of UTS. However, we could still observe challenging cases both in the real-world dataset and the synthetic dataset that indicate that UTS can be further improved in the future. Fig.~\ref{fig:experiments-qualitative-error-case} exhibits the dominant sources of error, which are detection errors, dynamic occlusion and orientation initialization errors. Although we found out that the UKF is quite robust to initial errors and is able to recover fast during tracking the initial errors still have an impact on the evaluation.

%% file: sourcefiles/conclusion.tex
\section{Conclusion \& Outlook}

We have developed a novel surveillance system designed for demanding urban traffic scenarios. UTS operates in a 3D space with measurements only coming from a 2D space. Due to the lack of prior work in this area, we utilized CARLA to generate synthetic datasets for the purpose of quantitatively evaluating our methods. This quantitative evaluation on synthetic data as well as a qualitative evaluation on real-world data allowed us to give a first impression of the overall very promising performance of UTS and also to identify the main challenges where our method does not yet perform optimally and can be further improved by future work.

The rarely but still regularly occurring detection errors that especially happen with very large vehicles as trucks and buses can be taken care of by an additional training of the 2D detector (which at the moment is a general detector not specialized for a wide range of vehicles) on images containing specifically these classes. Occlusion is another major problem. When a vehicle is only partly and temporarily occluded, then occlusion can be handled by identifying and disregarding the edges of the 2D detection box that are affected. UTS is already robust to these occlusions if they are caused by static objects. The system considers a manually created mask where areas are labeled that might perform static occlusions. The dynamic case can be taken care of in the future by exploiting the reconstructed 3D information of the observed vehicles before they occlude each other.

Nevertheless, permanent or long-continued occlusions continue to be a problem of surveillance systems that rely on only one sensor with a limited field of view. Obviously, this problem could be tackled by installing several cameras, each pointing at the same crossroad from a different perspective. This multi-sensor approach should not yield multiple observations of the same object but perform a sensor fusion. By that, every vehicle can be tracked accurately by at least one of the cameras and the tracking and 3D reconstruction can profit from the various perspectives of the camera network. This would also improve situations where 3D information is estimated for far field observations as a network of well distributed sensors would significantly reduce the maximum observation distance. Such a multi sensor system would then require a mechanism that associates and subsequently fuses the observations of all sensors and is part of our ongoing work.